\documentclass[10pt,twocolumn,letterpaper]{article}

\usepackage[pagenumbers]{cvpr} 
\usepackage{bbding}
\definecolor{cvprblue}{rgb}{0.21,0.49,0.74}

\definecolor{shadecolor}{rgb}{0.92, 0.92, 0.92}
\definecolor{gtgray}{gray}{0.97}
\definecolor{mygray}{gray}{.88}

\definecolor{gray1}{gray}{.90}
\definecolor{gray2}{gray}{.92}
\definecolor{gray3}{gray}{.94}

\usepackage{xcolor} 
\usepackage{colortbl}

\usepackage[pagebackref,breaklinks,colorlinks,allcolors=cvprblue]{hyperref}
\usepackage[most]{tcolorbox} 
\usepackage{multirow}

\usepackage{textcomp} 
\renewcommand{\thefootnote}{\fnsymbol{footnote}} 

\def\paperID{--} 
\def\confName{CVPR}
\def\confYear{2026}

\title{TDMM-LM: Bridging Facial Understanding and Animation via Language Models}
\author{\footnotemark[2]~~Luchuan Song$^{1}$, \footnotemark[1]~~Pinxin Liu$^{1}$, Haiyang Liu$^2$, Zhenchao Jin, \\ Yolo Yunlong Tang$^1$, Zichong Xu$^1$, Susan Liang$^1$, Jing Bi$^1$, Jason J Corso$^{3,4}$, Chenliang Xu$^1$\\
\vspace{-10pt}
\and
$^1$University of Rochester \quad $^2$University of Tokyo \quad $^3$University of Michigan \quad $^4$Voxel51\\
{\tt\small lsong11@ur.rochester.edu, \{pliu23, yunlong.tang, sliang22\}@ur.rochester.edu,} \\{\tt\small haiyangliu1997@gmail.com, jjcorso@eecs.umich.edu, chenliang.xu@rochester.edu}
}

\begin{document}

\twocolumn[{%
\renewcommand\twocolumn[1][]{#1}%
\maketitle
\begin{center}
    \centering
    \vspace{-.5cm}
    \captionsetup{type=figure}
    \includegraphics[width=1.\textwidth]{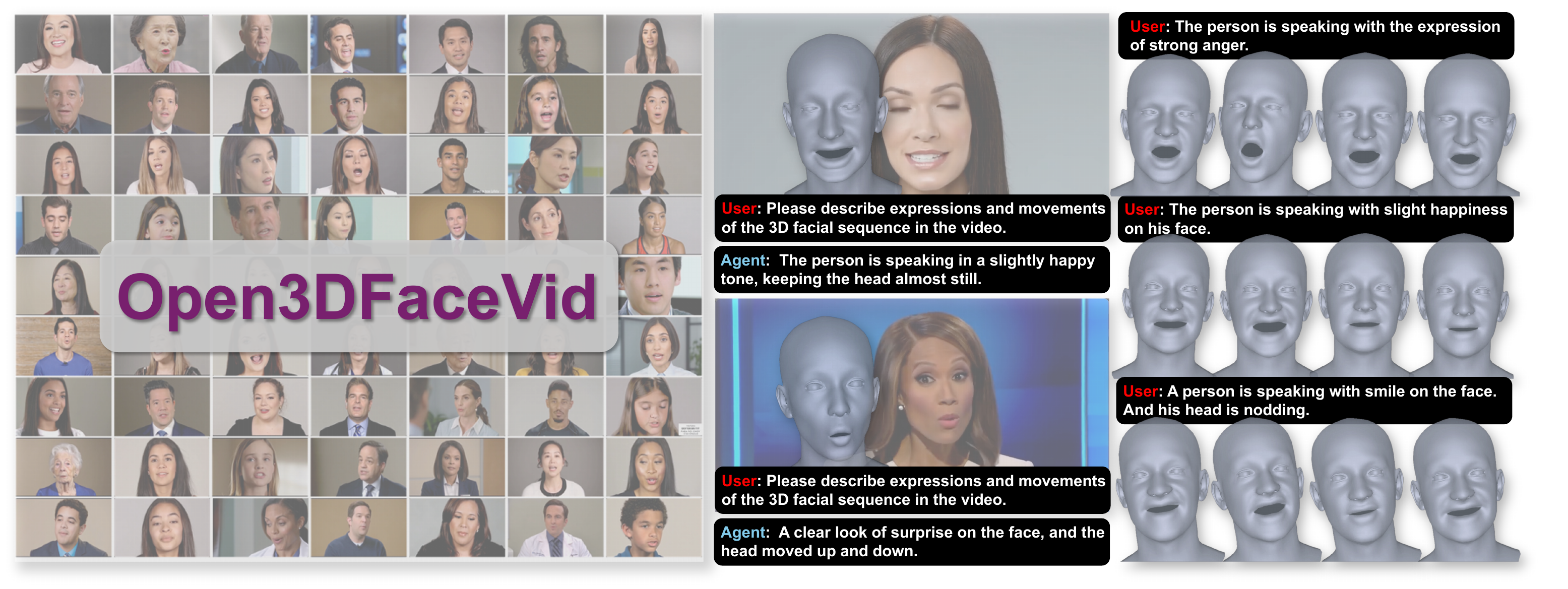}
    \vspace{-.8cm}
    \captionof{figure}{Overview of the proposed Open3DFaceVid dataset and 3D facial understanding/animation pipeline. The left panel visualizes the Open3DFaceVid corpus, which covers a wide range of identities, emotions, and speaking styles generated via text-to-video (T2V) models. The right panel illustrates our interactive 3D facial interface: given a 3DMM sequence, the user prompts the agent to describe expressions and head motion in natural language, and the agent returns fine-grained, parameter-based interpretations. In the reverse direction, the agent is able to condition on user prompts to generate new 3DMM trajectories with controllable emotion and pose. Please refer to \url{https://songluchuan.github.io/TDMM-LM/} for visualization results and datasets.}
\label{Teaser}
\end{center}%
}]

\makeatletter
\def\hlinew#1{%
  \noalign{\ifnum0=`}\fi\hrule \@height #1 \futurelet
   \reserved@a\@xhline}
\makeatother
\begingroup
\renewcommand{\thefootnote}{\fnsymbol{footnote}}
\footnotetext[2]{\ Project Leader.}
\footnotetext[1]{Equal contribution.}
\endgroup

\maketitle

\begin{abstract}
Text-guided human body animation has advanced rapidly, yet facial animation lags due to the scarcity of well-annotated, text-paired facial corpora. To close this gap, we leverage foundation generative models to synthesize a large, balanced corpus of facial behavior. We design prompts suite covering emotions and head motions, generate about 80 hours of facial videos with multiple generators, and fit per-frame 3D facial parameters, yielding large-scale (prompt and parameter) pairs for training. Building on this dataset, we probe language models for bidirectional competence over facial motion via two complementary tasks: (1) Motion2Language: given a sequence of 3D facial parameters, the model produces natural-language descriptions capturing content, style, and dynamics; and (2) Language2Motion: given a prompt, the model synthesizes the corresponding sequence of 3D facial parameters via quantized motion tokens for downstream animation. Extensive experiments show that in this setting language models can both interpret and synthesize facial motion with strong generalization. To best of our knowledge, this is the first work to cast facial-parameter modeling as a language problem, establishing a unified path for text-conditioned facial animation and motion understanding. 
\end{abstract}    
\vspace{-0.4cm}
\section{Introduction}
\label{sec:intro}

\begin{figure*}[htp]
\centering
\vspace{-0.4cm}
  \includegraphics[width=2\columnwidth, trim={0cm 0cm 0cm 0cm}, clip]{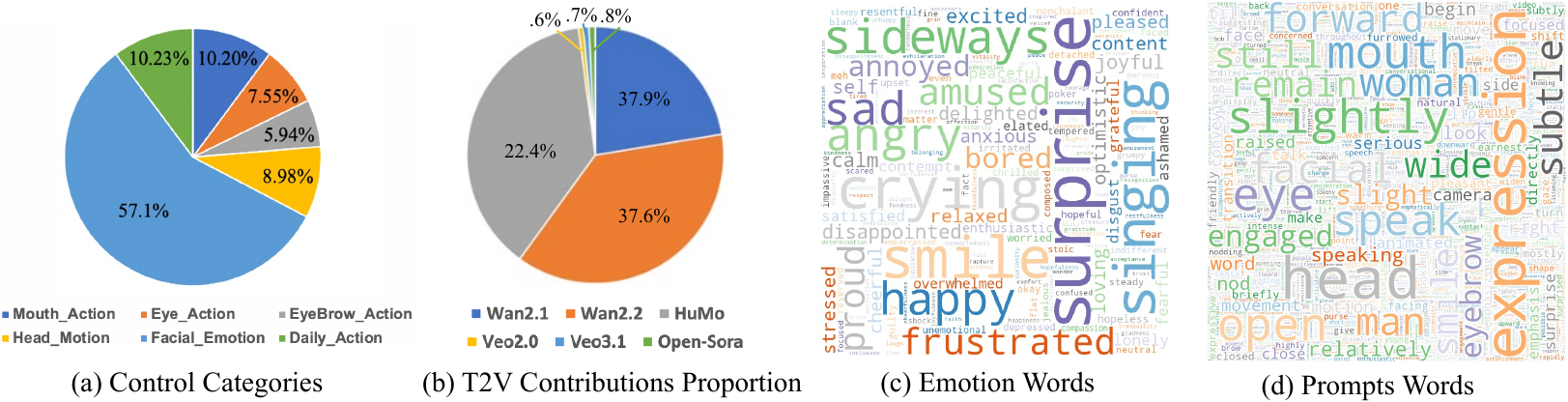}
  \vspace{-0.1cm}
\caption{The analysis of the Open3DFaceVid dataset. We summarize the control categories induced by prompts and their corresponding video counts, broken down by underlying T2V backbones. We further visualize the vocabulary with word clouds, separately for emotion-related terms and for full-text prompts, to highlight the diversity and saliency of affective descriptors.}
\label{fig:pipeline}
\vspace{-4mm}
\end{figure*}

Multimodal large language models (MLLMs)~\cite{yin2024survey,shen2024aligning,lu2024mace,zhang2022dino,li2023llava,lu2025does,yu2025visual,zhang2305pmc,lin2024video,sun2024video,chen2024sharegpt4video,huang2024vtimellm,shu2025audio,zheng2025video,huang2025directional}
have significantly advanced visual understanding through joint reasoning over images, audio, and language. 
Modern systems~\cite{team2023gemini,wang2024qwen2,bai2025qwen2,kong2024hunyuanvideo,lin2024moe} achieve state-of-the-art results on a wide range of perception tasks~\cite{wu2021towards,lu2022copy,lu2024robust,feichtenhofer2019slowfast,lin2019tsm,diba2020large}. 
Recent domain-specialized MLLMs~\cite{zhao2025humanomni,hong20233d,maaz2024video,sun2024hicmae,yang2025omni} further demonstrate strong potential for human-centered visual reasoning.

However, fine-grained facial-behavior understanding remains fundamentally limited. \textbf{\emph{Token inefficiency}} is a major bottleneck: existing V-LLMs must process every frame as a large set of image tokens. 
To reduce cost, models downsample video or sparsely sample keyframes. 
While acceptable for coarse actions, this is detrimental for facial motion, where expressions unfold over only a few frames and temporal continuity is crucial. 
Micro-expressions such as brow raises, lip twitches, or brief smirks often disappear when frames are dropped; even retained frames yield hundreds of image tokens per second, restricting temporal context and forcing models to ignore subtle dynamics. 
The result is a systematic bias toward static, neutral facial behavior. \textbf{\emph{Emotion imbalance}} in existing training corpora amplifies this issue. 
Large-scale MLLMs~\cite{team2023gemini,zhao2025humanomni,yang2025omni} are mostly trained on in-the-wild videos (\eg, YouTube, TikTok, VoxCeleb~\cite{nagrani2020voxceleb,chung2018voxceleb2}), which overwhelmingly feature neutral, frontal talking heads. 
High-intensity or atypical expressions (e.g. pouting, frowning, laughing, smirking) are rare. Coupled with temporal sparsity from downsampling, models learn a narrow, low-variance facial prior and struggle to perceive or generate expressive motion.

These limitations raise a central question:  
``\textit{Can fine-grained facial emotion and movement be represented using low-dimensional geometric signals (e.g., 3D facial parameters) that preserve subtle temporal variation while avoiding redundant visual tokens?}" A major obstacle is the lack of class-balanced, richly annotated facial-motion data. 
Existing datasets~\cite{wang2020mead,harte2015tcd,livingstone2018ryerson,czyzewski2017audio,huang2023egocentric} rely on studio capture and provide coarse, one-hot emotion labels, lacking the diversity and open-form descriptions needed for language-driven modeling.
To overcome this limitation, we introduce \textbf{Open3DFaceVid}, a scalable synthetic pipeline that generates diverse facial videos using multiple T2V models~\cite{wan2025wan,wiedemer2025video}. 
We curate a lexicon of $\sim$200 emotion and facial-action descriptors (\eg, grin, pout, smirk, squint), uniformly sample all categories to ensure balanced coverage, and extract 3DMM parameters~\cite{egger20203d} for every clip. 
This yields a large corpus of paired videos, text descriptions, and high-quality 3D facial-motion trajectories. 
Dataset characteristics are shown in Fig.~\ref{fig:pipeline}.

With this dataset providing the necessary coverage and annotation richness, we explore a modeling paradigm that \emph{bypasses image tokens entirely}. We train a geometry VQ-VAE~\cite{van2017neural} that discretizes 3DMM sequences into a perceptually coherent codebook, producing compact geometry tokens that replace image tokens in the MLLM input space. This structured representation preserves subtle dynamics, removes visual redundancy, and drastically reduces token consumption. 
Finetuning a LLM on paired motion–text samples enables it to directly reason over 3D facial motion.

Built on this alignment interface, our framework supports two complementary directions using the \textit{same} LLM: (1) \textbf{Motion2Language}: interpreting geometry-token sequences to produce natural-language descriptions of emotion, intensity, micro-expressions, and head motion; (2) \textbf{Language2Motion}: generating expressive 3D facial-motion trajectories directly from user prompts, enabling fine-grained, text-driven facial animation.
Extensive experiments validate the effectiveness of our dataset, geometry-token representation, and unified modeling framework. 
We demonstrate strong performance across facial-motion understanding and language-driven animation. In summary, our contributions include:

\begin{itemize}
\item We present \textbf{Open3DFaceVid}, an $\sim$80-hour synthetic facial-motion corpus generated using multiple foundation T2V models. It provides the largest collection to date of paired Text2Face annotations with diverse emotions, subjects, and intensity variations. \par

\item We extend the LLM to the \textbf{Motion2Language} setting, enabling natural-language interpretation of 3D facial motion. Given a facial-motion token sequence, the model produces compact, expressive descriptions of emotions, micro-expressions, and head movements. \par

\item We further introduce a \textbf{Language2Motion} framework that generates 3D facial motion from natural-language prompts. By conditioning an autoregressive geometry decoder on word-level LLM embeddings, users can precisely control fine-grained facial dynamics through text. \par

\end{itemize}

\section{Related Work}
\label{sec:related_work}

\begin{figure*}[htp]
\centering
\vspace{-0.2cm}
  \includegraphics[width=2.\columnwidth, trim={0cm 0cm 0cm 0cm}, clip]{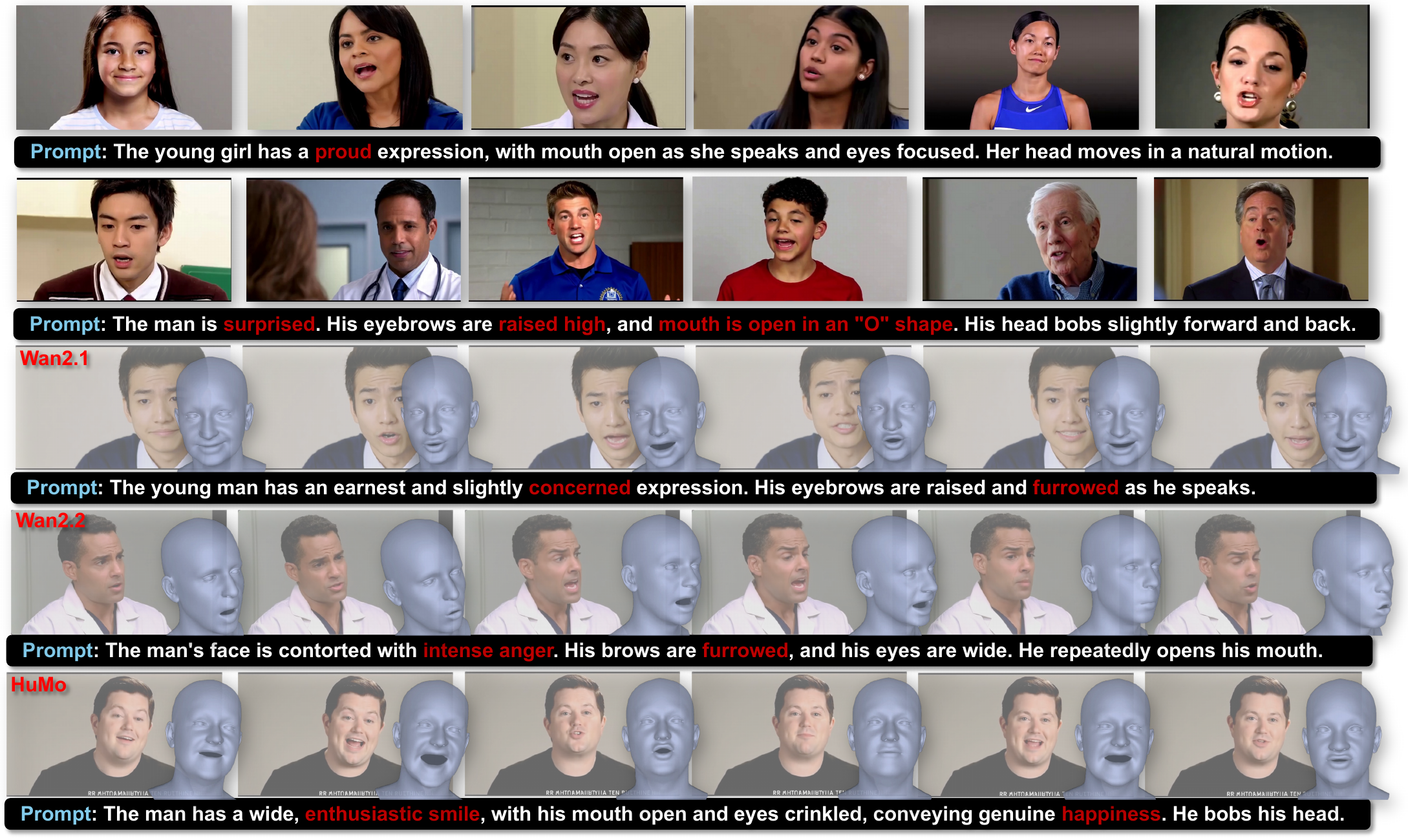}
  \vspace{-0.4cm}
\caption{\textbf{Dataset overview.} Top two rows: starting from a fixed text prompt, we vary the random seed and emphasize different prompt keywords to modulate facial identity and video attributes, showcasing subjects across different genders. Bottom three rows: we recover FLAME facial parameters and pair the resulting trajectories with the corresponding prompt, forming Text–3DMM dataset.}
\label{fig:pipeline}
\vspace{-2mm}
\end{figure*}

\noindent \textbf{Text-to-Motion Generation}
Text–motion alignment underpins much of modern motion understanding~\cite{oord2019representationlearningcontrastivepredictive,liu2025contextualgesturecospeechgesture,liu2025intentionalgesturedeliverintentions}, with models such as MotionCLIP~\cite{tevet2022motionclip} and TMR~\cite{petrovich2023tmr} mapping language descriptions to dynamic motion sequences. Building on this foundation, recent work has shifted toward text-to-motion generation: autoregressive models~\cite{zhang2023generating,ribeiro2024motiongpt, guo2022generating, lu2023humantomato, guo2024momask,kinmo2025kinematicawarehumanmotion} tokenize motion and decode it in a language-like space. Supported by large-scale body motion datasets~\cite{plappert2016kit,punnakkal2021babel,guo2022generating,lin2023motion}, these approaches enable natural and expressive full-body movement generation. In contrast, progress in facial animation has lagged behind, largely due to the absence of emotion-rich, text-aligned corpora. 

\vspace{0.1cm}
\noindent \textbf{Text-to-Video Generation}
Text-to-video generation has made significant progress in recent years. Early diffusion-based~\cite{zhu2024oftsr,gao2024eraseanything,li2025set,zhou2025dragflow,zhao2026luve} frameworks factorize the generation via intermediate image conditioning, improving stability and quality~\cite{singer2024video}. More recent works target human-centric video or long shot sequences: Identity-Preserving T2V ensures consistent human identity while generating high-fidelity video~\cite{Yuan_2025_CVPR}, and Text-to-Multi-Shot Video Generation addresses transitions and scene continuity~\cite{kara2025shotadaptertexttomultishotvideogeneration}. Commercial systems such as Sora2~\cite{openai2024video}/Veo3~\cite{wiedemer2025video} have further demonstrated text-driven open-world video generation.

\section{Open3DFaceVid}
\label{sec:method}

\begin{figure}[]
\centering
  \includegraphics[width=.8\columnwidth, trim={0cm 0cm 0cm 0cm}, clip]{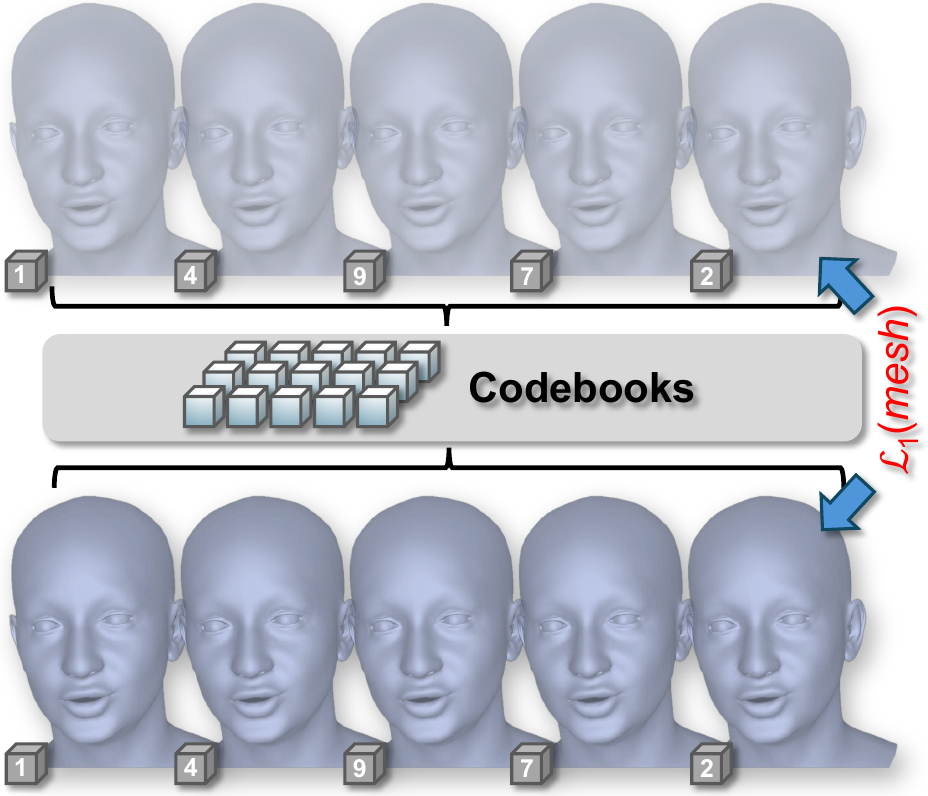}
\vspace{-2mm}
\caption{Geometry-aware facial tokenization learning. We quantize facial expression codes into a discrete codebook and enforce reconstruction in mesh space. The input facial expression codes are mapped to code indices, decoded back to FLAME meshes (bottom), and supervised with an $\mathcal{L}_1$ loss on vertex positions}
\vspace{-0.5cm}
\label{fig:pipeline-vq}
\end{figure}

We build our dataset by programmatically prompting a family of Text-to-Video (T2V) models~\cite{sun2024sora} to synthesize facial videos that can be reconstructed in our 3D setting. The prompt space covers a large attribute lexicon, including subject descriptors (\eg, male, female, young lady) and affective cues (\eg, joy, anger, smile), to encourage broad coverage over identities and emotions. To avoid overfitting to the biases of a single generator, we instantiate the pipeline with multiple underlying T2V models (\eg, Wan-2.1/2.2~\cite{wan2025wan}, Open-Sora~\cite{opensora}, HuMo~\cite{chen2025humo}, Veos~\cite{wiedemer2025video}), which expands the range of styles and mitigates model-specific artifacts. 

\vspace{0.1cm}
\noindent \textbf{Prompts Construction}
\label{Prompts Construction}
The prompt design factorizes into facial appearance and video dynamics. On the facial side, we sample a rich space of affective descriptions, varying both emotion type and intensity, and augment them with frequent micro-expressions such as blinking, pouting, and related facial actions to increase behavioral diversity. On the video side, we employ a small set of templated prompts that constrain camera motion, encouraging stable framing. For example, instructions like ``\textit{The frame remains steady, with the head and shoulders centered}" are applied to keep the subject fixed in view and reduce cinematic jitter. We try to maintain the balance between the number of facial emotion categories and the number of subject categories in the prompts. Programmatically generated templates are then refined with the ChatGPT-4.1~\cite{achiam2023gpt}, yielding natural, conversational descriptions that better resemble daily language while preserving the intended control condition. 

\begin{figure}[]
\vspace{-5mm}
\centering
  \includegraphics[width=1\columnwidth, trim={0cm 0cm 0cm 0cm}, clip]{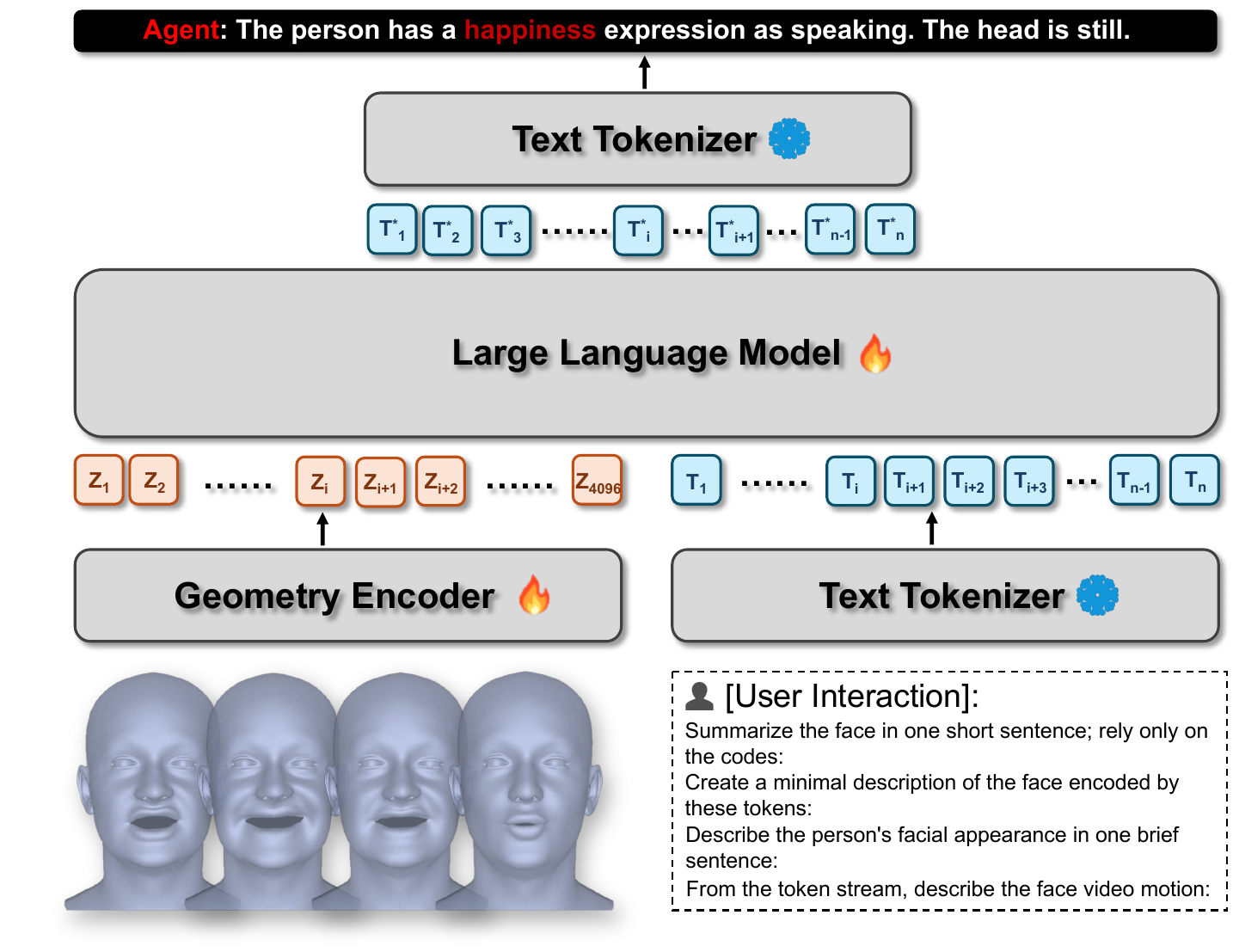}
\vspace{-0.7cm}
\caption{\textbf{Motion2Language.} Geometry sequences are encoded into discrete facial tokens by the geometry encoder and fed, together with text tokens from the user prompt, into a LLM. Conditioned only on these geometry tokens, the agent generates natural-language descriptions of expression/head motion, enabling interactive question to answering about 3D facial behavior.}
\vspace{-0.3cm}
\label{fig:pipeline-llm}
\end{figure}

\vspace{0.1cm}
\noindent \textbf{Human-Centric T2V Synthesis}
\label{Human-Centric T2V Synthesis}
We apply the polished prompts as control condition to synthesize a large collection of facial videos. To avoid collapsing the motion distribution onto the prior of any single generator, we instantiate the pipeline with a suite of base T2V models. For each backbone, the facial-attribute portion of the prompt is kept standardized, while only the video-related clauses are lightly adapted to match model-specific preferences in framing and cinematography. Our corpus contains roughly 60K synthetic clips, each about 4–6 seconds. To narrow the gap to real data, we augment this set with about 10K wild clips and derive prompt-style annotations with a vision–language model (Gemini~\cite{team2023gemini}). 

\vspace{0.1cm}
\noindent \textbf{Geometry Facial Estimation}
\label{3D Facial Estimation}
We estimate facial motion with 3D Morphable Model (3DMM)~\cite{egger20203d} parameters, fitted to monocular videos in our corpus to recover per-clip identity, expression, and head pose. Specifically, we adopt the FLAME~\cite{li2017learning} model to regress facial parameters.

\begin{figure}[]
\vspace{-5mm}
\centering
  \includegraphics[width=1\columnwidth, trim={0cm 0cm 0cm 0cm}, clip]{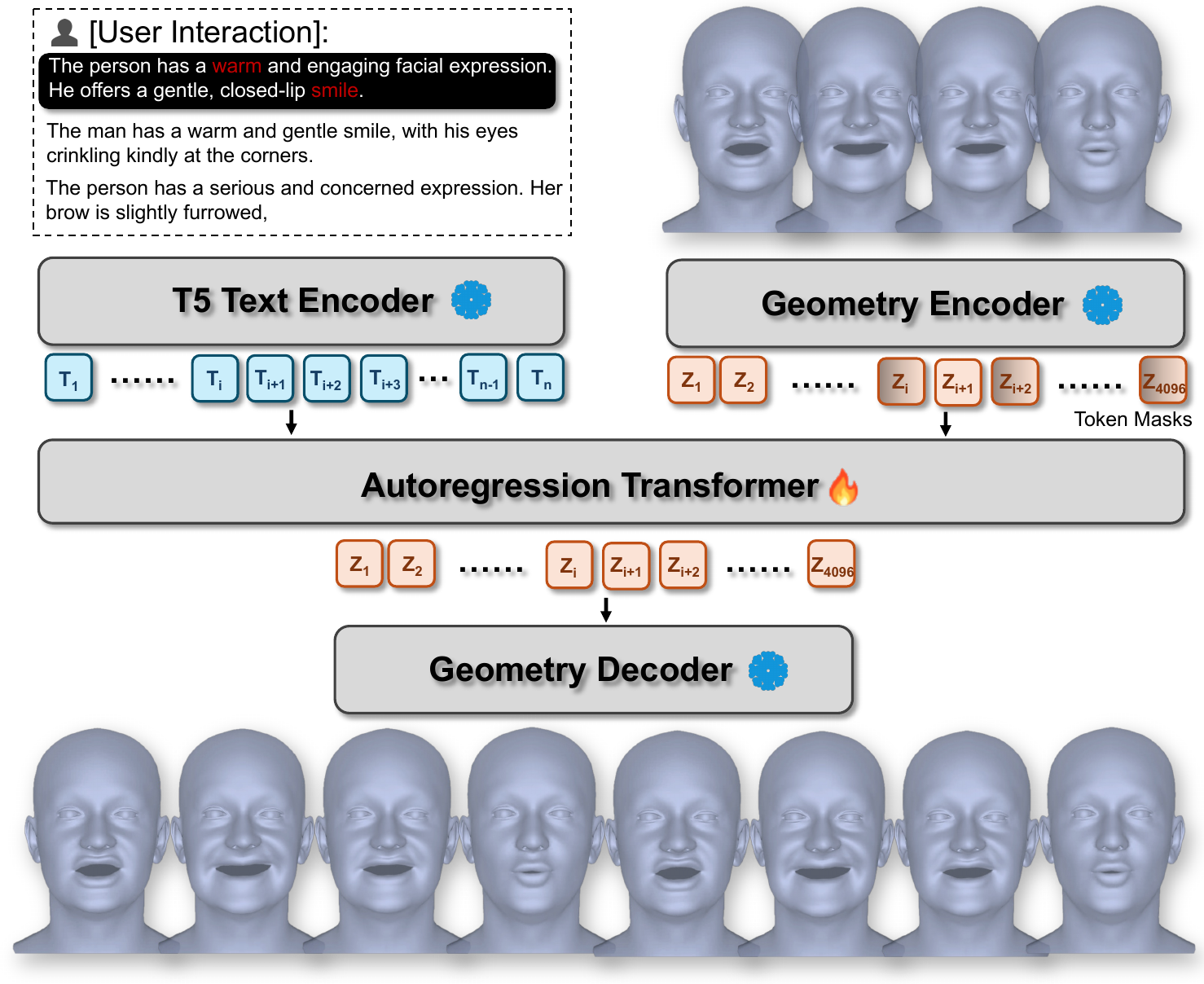}
\vspace{-0.7cm}
\caption{\textbf{Language2Motion.} The user provides a natural-language description of the desired facial behavior (top left). The text tokenizer converts the prompt into word-level tokens, while a paired 3D facial sequence is encoded into discrete geometry tokens by the geometry encoder. The autoregressive transformer predicts future geometry tokens conditioned on the text prefix.}
\vspace{-0.1cm}
\label{fig:pipeline-l2m}
\end{figure}

\section{Language-Motion Alignment}
\label{Language-Motion Alignment}

Base on \textbf{Open3DFaceVid}, we learn a shared space between 3D facial motion (3DMM) and natural language, enabling both interpretation and synthesis of expressive facial behavior. The alignment pipeline has three key components: (1) the \textit{Geometry VQ-VAE} operates on facial geometry rather than continuous 3DMM parameters, mapping sequences into the discrete token space; 
(2) an \textit{LLM-based motion interpreter} that directly decodes geometry-token sequences into natural-language descriptions, supporting Motion2Language understanding;  
(3) a \textit{shared language–motion transformer} that conditions geometry-token prediction on word-level embeddings, enabling controllable Language2Motion synthesis.

\vspace{0.1cm}
\noindent\textbf{Geometry VQ-VAE}
Shown in Fig.~\ref{fig:pipeline-vq}, we quantize facial dynamics into a discrete latent space. Rather than discretizing on 3DMM parametersm which can map similar expressions to different coefficient patterns. We operate directly on reconstructed facial geometry, ensuring that visually similar expressions are encoded with consistent tokens. This design mitigates the many-to-one ambiguity in which multiple expression codes map to nearly facial geometry. 

\vspace{0.1cm}
\noindent \textbf{Motion2Language} 
To translate 3D facial motion into natural language, we utilize geometry tokenizer to tokenize facial geometry, where discrete tokens are fed as symbolic observations to the LLM. Instead of relying on image inputs, the LLM directly consumes sequences of geometry tokens, producing rich free-form descriptions of emotions, intensities, micro-expressions (e.g., blinking, pouting), and head dynamics. Each training instance provides a geometry-token sequence and an accompanying textual description. To improve linguistic coverage without altering semantics, we expand each annotation with multiple paraphrases produced by an auxiliary LLM. As shown in Fig.~\ref{fig:pipeline-llm}, the LLM is instruction-tuned to map geometry-token observations to natural-language responses conditioned on task templates.

\vspace{0.1cm}
\noindent \textbf{Language2Motion} 
Compared to full-body scenarios, text-driven facial motion generation remains considerably more challenging. Available datasets are smaller, facial actions are subtler, and conventional text-to-motion pipelines compress the entire prompt into a global embedding, losing the token-level cues that dictate fine-grained muscular dynamics. To leverage the  \emph{linguistic granularity}, we introduce a \emph{word-level language prefix} that injects pretrained LLM embeddings into an autoregressive facial-motion transformer. Importantly, this prefix is processed by the text encoder which takes the user's text prompt, produces token-level embeddings, and conditions the motion decoder without modifying the vocabulary of the language model. As illustrated in Fig.~\ref{fig:pipeline-l2m}, this prefix-based fusion preserves the structure of the prompt and permits individual words to steer localized facial movements, enabling controllable and semantically aligned 3D facial-motion generation.

\section{Experiments}
\label{sec:experiments}

\label{Implementation Details}

\begin{table}[t]
\footnotesize
\begin{center}
\vspace{-0.2cm}
\setlength{\tabcolsep}{.6mm}
{
\begin{tabular}{c|ccc|cc}
\hlinew{.8pt}
\multirow{2}{*}{Methods} &$\text{Cor}_{E}$$\uparrow$ &$\text{Cor}_{M}$$\uparrow$&$\text{Cor}_{I}$$\uparrow$&$\text{USER}_{E}$$\uparrow$&$\text{USER}_{M}$$\uparrow$\\
&\multicolumn{3}{c|}{[GPT-4]}  &  \multicolumn{2}{c}{[Human]} \\
\cline{2-6}
\hline
HumanOmni~\cite{sun2024diffposetalk} & 1.84 & 1.17 & 1.09  & 2.04 & 1.00   \\
Gemini2.5 VLM~\cite{wang2020mead} &2.45 & 2.91 & 3.51 & 2.88 & 3.41    \\
\textbf{Ours} &\textbf{4.02} & \textbf{3.35} &\textbf{3.63} &\textbf{4.29}  &\textbf{3.79}  \\
\hlinew{.8pt}
\end{tabular}}
\vspace{-0.3cm}
\caption{Quantitative evaluation of Motion2Language. Mean 1–5 correctness scores from GPT-4 and human raters for emotion ($\text{Cor}_{E}$, $\text{USER}_{E}$), motion ($\text{Cor}_{M}$, $\text{USER}_{M}$) and intensity ($\text{Cor}_{I}$). We bold the best. Our geometry-token model consistently outperforms HumanOmni and Gemini-2.5 VLM across the metrics.}
\vspace{-0.7cm}
\label{table_2}
\end{center}
\end{table}

\begin{table}[!t]
\footnotesize
\begin{center}
\vspace{-0.4cm}
\setlength{\tabcolsep}{.6mm}
{
\begin{tabular}{c|ccc|cc}
\hlinew{.8pt}
\multirow{2}{*}{Methods} &$\text{Cor}_{E}$$\uparrow$ &$\text{Cor}_{M}$$\uparrow$&$\text{Cor}_{I}$$\uparrow$&$\text{USER}_{E}$$\uparrow$&$\text{USER}_{M}$$\uparrow$\\
&\multicolumn{3}{c|}{[GPT-4]}  &  \multicolumn{2}{c}{[Human]} \\
\cline{2-6}
\hline
HumanOmni~\cite{zhao2025humanomni} & 3.66 & 1.59 & 1.42  & 3.82 & 1.27   \\
Gemini2.5 VLM~\cite{team2023gemini} &\textbf{4.21} & 3.17 & \textbf{3.92} & 4.17 & \textbf{3.82}    \\
\textbf{Ours} &4.02 & \textbf{3.35} &3.63 &\textbf{4.29}  &3.79  \\
\hlinew{.8pt}
\end{tabular}}
\vspace{-0.3cm}
\caption{Quantitative evaluation of Motion2Language with natural images input. We employ the same correctness metric as in Table~\ref{table_2}, while replacing geometry-image inputs with natural facial images to better adapt the evaluation setting to VLMs.}
\vspace{-.5cm}
\label{table_rebuttal}
\end{center}
\end{table}

\subsection{Dataset Details} 

The Open3DFaceVid has a total duration of 81 hours and contains 57.2K video clips ranging from 4 to 6 seconds. The videos in datasets are standardized to 25 FPS and the resolution is resized $618 \times 360$. We employ 32 H200 GPUs to synthesize videos in T2V setting using Wan2.2 (5B and 14B), HuMo (17B), Open-Sora (11B)  and Wan2.1 (14B), consuming roughly 400 GPU Hours in total. We obtain Veo2/3 samples via its API interface. Due to the high per-clip generation cost, we include a modest number of Veo2/3 generated videos in our corpus. For face estimation, we use the FLAME template to regress per-frame expression, and head rotation (yaw/roll/pitch) parameters within videos. 
We present the model implementation details in the Appendix.

\begin{table}[t]
\footnotesize
\begin{center}
\setlength{\tabcolsep}{2.mm}
{
\begin{tabular}{c|ccc|cc}
\hlinew{.8pt}
\multirow{2}{*}{Methods} & $L_2$$\downarrow$ &FD$\downarrow$ &Tok$\uparrow$ &$\text{Cor}_{E}$$\uparrow$ & USER$\uparrow$\\
&\multicolumn{3}{c|}{[Parameters]}  &  \multicolumn{2}{c}{[Language]} \\
\cline{2-6}
\hline
T2M-X~\cite{liu2024t2mxlearningexpressivetexttomotion} & 0.471 & 47.59 & 0.671  & 2.21 & 3.40   \\
T2M-GPT~\cite{zhang2023generating} &0.226 & 37.04 & 0.895 & 3.57 & 3.91    \\
\textbf{Ours} &\textbf{0.219} & \textbf{31.75} &\textbf{0.920} &\textbf{4.13}  &\textbf{3.95}  \\
\hlinew{.8pt}
\end{tabular}}
\vspace{-0.2cm}
\caption{Quantitative evaluation of Language2Motion. Comparison with T2M-X and T2M-GPT on parameter-space measures ($L_2$/FD/Tok) and language-based metrics ($\text{Cor}_E$/USER). $L_2$/FD assess expression/pose fidelity, Tok is token-level accuracy.}
\vspace{-0.3cm}
\label{table_3}
\end{center}
\end{table}

\begin{figure}[]
\centering
  \includegraphics[width=1\columnwidth, trim={0cm 0cm 0cm 0cm}, clip]{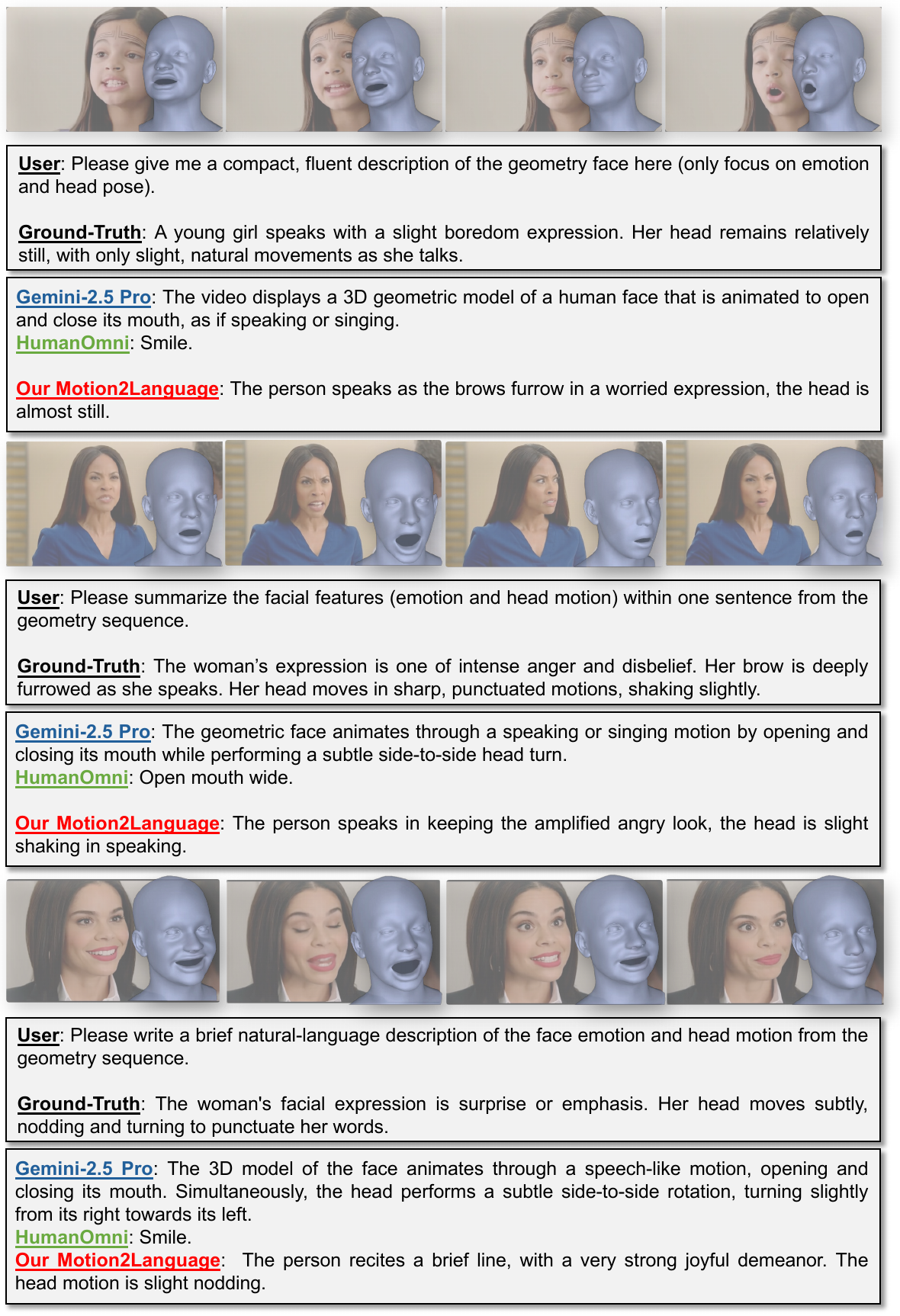}
\vspace{-0.7cm}
\caption{Qualitative comparison for Motion2Language. For representative clips, we show the input frames/geometry. The Gemini/HumanOmni operate on rendered geometry, while our model relies on 3D facial tokens. Our approach produces more accurate descriptions of emotion/motion, closely matching GT annotations.}
\vspace{-0.1cm}
\label{fig:pipeline-l2m-result}
\end{figure}

\subsection{Baselines} 
\label{Baseline}

Given the challenges of constructing text–motion datasets, there exist few directly comparable prior methods. We therefore adopt the most relevant techniques as baselines across both Motion2Language and Language2Motion.

\noindent \textbf{Motion2Language.}  
We evaluate against strong visual–language models applied directly to video frames.  
(1)~\textbf{HumanOmni~\cite{zhao2025humanomni}}, a human-centric VLM designed for holistic audio–visual reasoning. We treat it as a vision–only model by sampling video frames from dataset and prompting it to describe head motion and emotion. During inference, it operates entirely in pixel space.  
(2)~\textbf{Gemini VLM}, a general-purpose multimodal model~\cite{team2023gemini}. We uniformly sample video frames and provide an instruction prompt asking for descriptions of head pose and affective state.

\noindent \textbf{Language2Motion.}  
We draw baselines from text-to-human-motion generation.  
(3)~\textbf{T2M-X~\cite{liu2024t2mxlearningexpressivetexttomotion}}, a transformer-based text-to-motion model that represents motion as discrete tokens and autoregressively predicts pose sequences; we adapt its facial-motion branch as a baseline.  
(4)~\textbf{T2M-GPT~\cite{zhang2023generating}}, a GPT-style generative model trained on discrete motion codes. It directly models the distribution of tokenized motion trajectories conditioned on text descriptions. We follow their discrete VQ-token formulation and train the facial-motion pathway on our dataset for fair comparison.

\begin{figure}[]
\centering
  \includegraphics[width=1\columnwidth, trim={0cm 0cm 0cm 0cm}, clip]{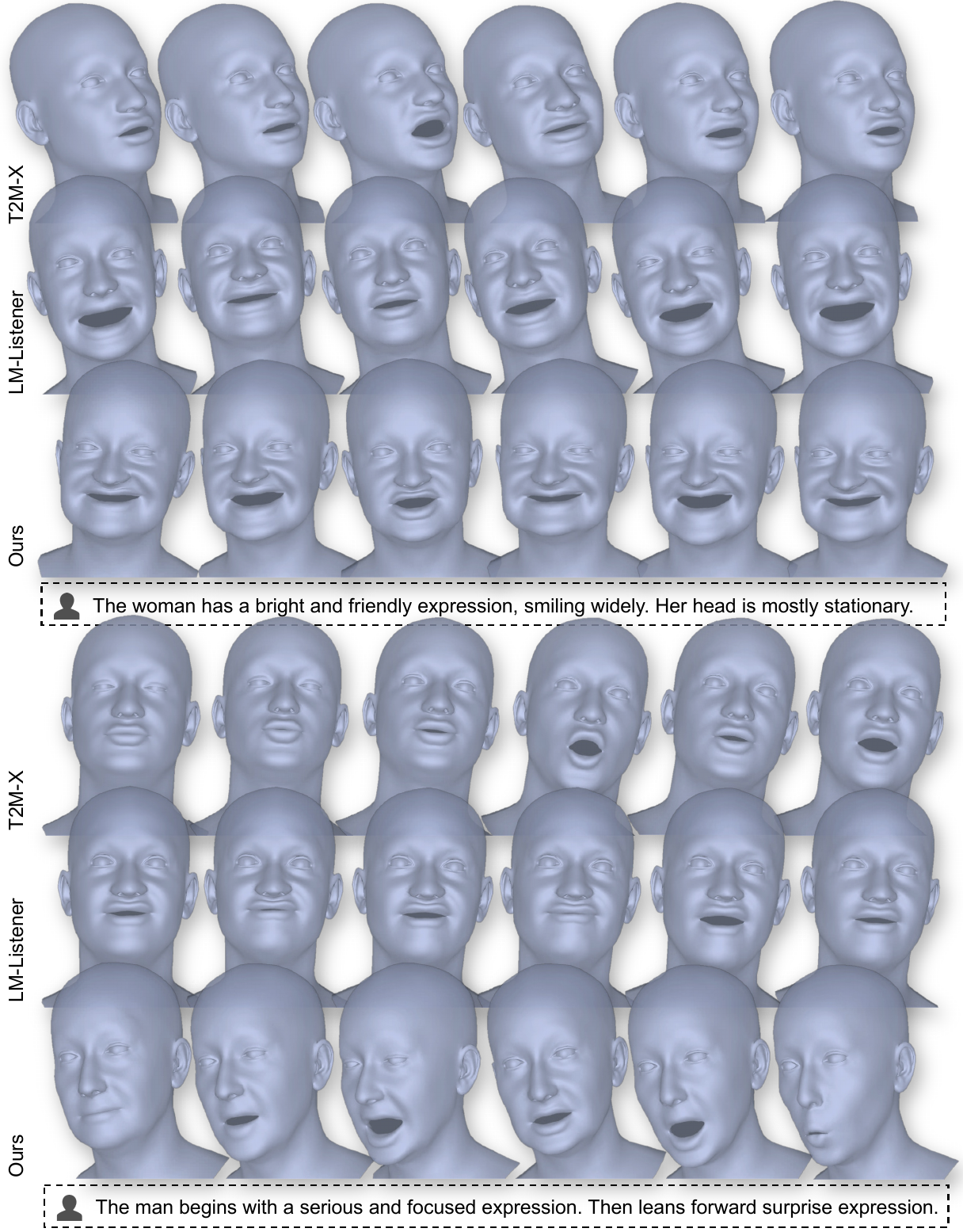}
\vspace{-3mm}
\vspace{-0.3cm}
\caption{Qualitative comparison for Language2Motion. Given prompts, we visualize generated 3D facial motion from different method. Our model produces expressions and head poses that more faithfully follow the described affect}
\vspace{-0.3cm}
\label{fig:pipeline-l2m-result-1}
\end{figure}

\begin{figure*}[htp]
\vspace{-0.2cm}
\centering
  \includegraphics[width=2.\columnwidth, trim={0cm 0cm 0cm 0cm}, clip]{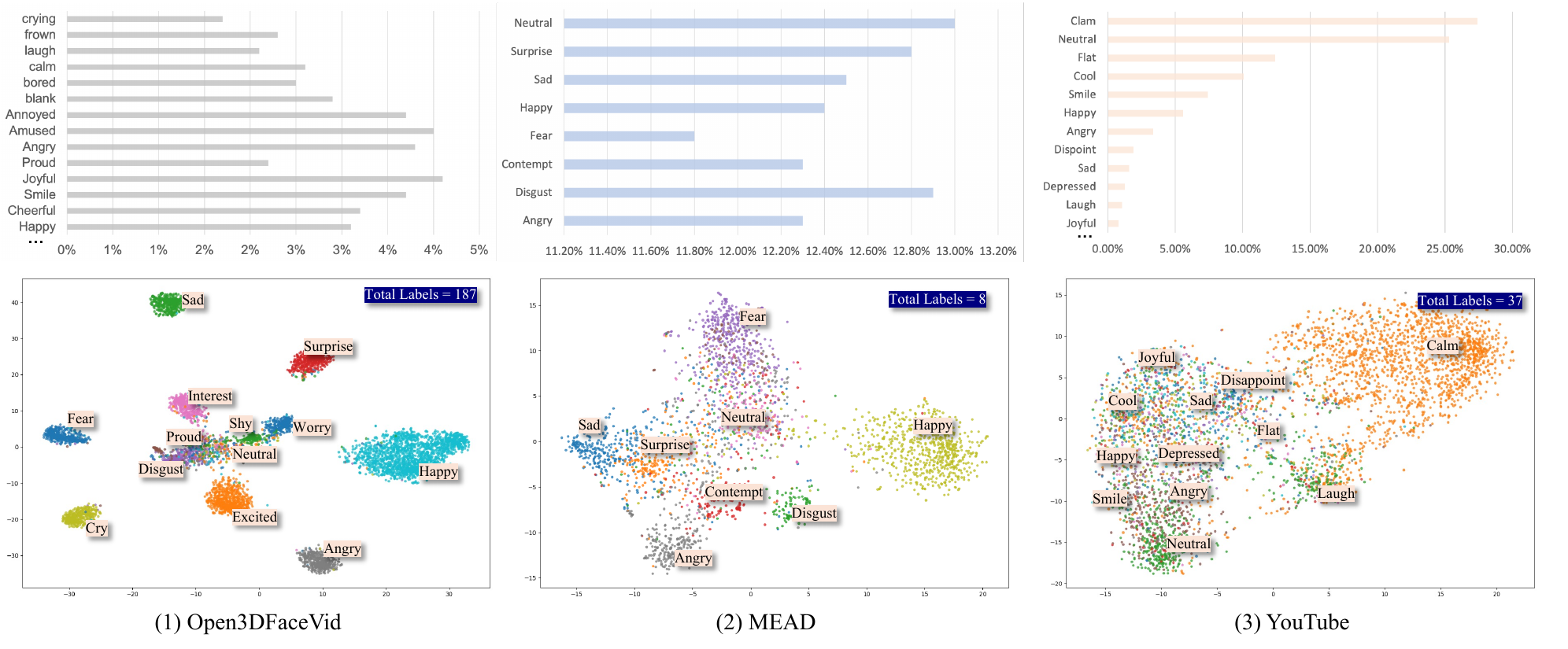}
  \vspace{-0.2cm}
\caption{\textbf{Emotion coverage across datasets.} Top-row: emotion-label distributions for (1) Open3DFaceVid, (2) MEAD~\cite{wang2020mead}, and (3) YouTube-derived clips, showing the number and balance of emotion categories (187/8/37 labels, respectively). Bottom-row: 2D projections of the corresponding facial (expression+pose) video embeddings, where Open3DFaceVid exhibits rich, well-separated clusters, while MEAD and YouTube provide coarser and less diverse affective coverage. Each point in the t-SNE~\cite{maaten2008visualizing} plots corresponds to the facial-expression features of a single video clip. Due to color variety limitations, we are unable to display all categories.}
\label{fig:tsne}
\end{figure*}

\subsection{Quantitative and Qualitative Evaluations}

\label{Quant./Qual. on Motion2Language}

\noindent\textbf{Motion2Language.} We validate:  
(1) geometry-token sequences preserve sufficient expressive information for accurate semantic interpretation, and  
(2) the compact tokenization provides higher efficiency than image-token VLMs.  
To quantify these aspects, we measure correctness in expression ($\text{Cor}_{E}$), motion ($\text{Cor}_{M}$), and intensity ($\text{Cor}_{I}$) using GPT-4 as an automatic judge on the 2K test clips, and collect human ratings ($\text{USER}_{E}$, $\text{USER}_{M}$) on 300 samples. Moreover, to ensure a fair comparison with the baseline methods, we use natural facial videos/images as inputs, as shown in Tab.~\ref{table_rebuttal}. We observe that commercial models such as Gemini achieve leading performance in the natural-image setting. Nevertheless, it is important to note that our method uses only a single token as input for each image, which is substantially fewer than the number of input tokens required by Gemini. 

Fig.~\ref{fig:pipeline-l2m-result} compares model outputs under identical prompts. HumanOmni and Gemini-2.5 Pro, both trained primarily on natural images, struggle to interpret temporal dynamics of facial motions, resulting in correctness scores close to 1. This reflects a domain gap rather than a failure of the task itself. In contrast, our model consistently recovers the correct emotion and motion semantics from 3DMM-derived tokens, demonstrating that structured geometric representations retain the key information needed for facial-behavior reasoning. Furthermore, our method encodes each frame with a single geometry token instead of 300–500 visual tokens, yielding a more efficient and temporally responsive understanding pipeline.

\begin{table}[t]
\footnotesize
\begin{center}
\setlength{\tabcolsep}{1.2mm}
{
\begin{tabular}{c|ccc|cc}
\hlinew{.8pt}
\multirow{2}{*}{Train-Corpus} &$\text{Cor}_{E}$$\uparrow$ &$\text{Cor}_{M}$$\uparrow$&$\text{Cor}_{I}$$\uparrow$&$\text{USER}_{E}$$\uparrow$&$\text{USER}_{M}$$\uparrow$\\
&\multicolumn{3}{c|}{[GPT-4]}  &  \multicolumn{2}{c}{[Human]} \\
\cline{2-6}
\hline
MEAD~\cite{wang2020mead} & 3.03 & 1.29 & 3.59  & 2.92 & 1.18   \\
YouTube [Gemini] &2.74 & \textbf{3.92} & 3.76 & 3.14 & 3.65    \\
\textbf{Open3DFaceVid} &\textbf{3.97} & {3.44} &\textbf{3.85} &\textbf{4.05}  &\textbf{3.75}  \\
\hlinew{.8pt}
\end{tabular}}
\vspace{-0.2cm}
\caption{Ablation on training corpus for Motion2Language. Identical architectures are trained on MEAD, YouTube (Gemini-annotated) and Open3DFaceVid. The metrics defined in Sec.~\ref{Quant./Qual. on Motion2Language}.}
\vspace{-0.5cm}
\label{table_4}
\end{center}
\end{table}

\begin{table}[t]
\footnotesize
\begin{center}
\vspace{-0.1cm}
\setlength{\tabcolsep}{2.mm}
{
\begin{tabular}{c|ccc|cc}
\hlinew{.8pt}
\multirow{2}{*}{Train-Corpus} & $L_2$$\downarrow$ &FD$\downarrow$ &Tok$\uparrow$ &$\text{Cor}_{E}$$\uparrow$ & USER$\uparrow$\\
&\multicolumn{3}{c|}{[Parameters]}  &  \multicolumn{2}{c}{[Language]} \\
\cline{2-6}
\hline
MEAD~\cite{wang2020mead} & \textbf{0.217} & \textbf{32.14} & 0.892  & 3.56 & 2.88   \\
YouTube [Gemini] &0.259 & 33.47 & 0.910 & 3.17 & 3.59    \\
\textbf{Open3DFaceVid} &{0.226} & {32.48} &\textbf{0.915} &\textbf{3.90}  &\textbf{3.92}  \\
\hlinew{.8pt}
\end{tabular}}
\vspace{-0.2cm}
\caption{Ablation on training corpus for Language2Motion. Identical models are trained on MEAD, YouTube (Gemini-annotated), and Open3DFaceVid. Training on Open3DFaceVid yields the best trade-off between geometric fidelity and text–motion alignment.}
\vspace{-.8cm}
\label{table_5}
\end{center}
\end{table}

\vspace{0.1cm}
\noindent\textbf{Language2Motion.} We test: (1) geometry-token generation conditioned on language can faithfully follow the intended facial semantics, and (2) the predicted 3DMM trajectories exhibit realistic temporal dynamics despite using a compact discrete representation.  
To validate these aspects, we adopt quantitative and human-centric metrics. Text–motion consistency is assessed through human ratings (USER) on 300 samples. 
Low-level reconstruction fidelity is measured using $L_2$ distance on expression and pose. 
Motion realism is evaluated via Fréchet Distance (FD), comparing feature distributions of generated and ground-truth sequences. 
Token prediction accuracy (Tok) reflects discrete code fidelity within our VQ token space. 
Finally, we compute expression correctness ($\text{Cor}_{E}$) using the trained Motion2Language model as an external semantic evaluator, providing an independent measure of alignment.

As shown in Tab.~\ref{table_3}, our method achieves consistently higher scores across all metrics, supporting both claims: the model generates semantically aligned facial motion and maintains high-quality temporal structure. 
T2M-GPT performs reasonably well due to its autoregressive architecture but falls short on semantic metrics, highlighting the benefit of conditioning on a stronger LLM backbone. 
Qualitative examples in Fig.~\ref{fig:pipeline-l2m-result-1} show that our model responds sensitively to nuanced instructions (\eg, transitioning from \textit{``focused''} to \textit{``surprised''}) and that higher-capacity variants capture expressive affect (\eg, \textit{``joyful''}) more distinctly than competing methods.

\begin{figure}[]
\centering
  \includegraphics[width=1\columnwidth, trim={0cm 0cm 0cm 0cm}, clip]{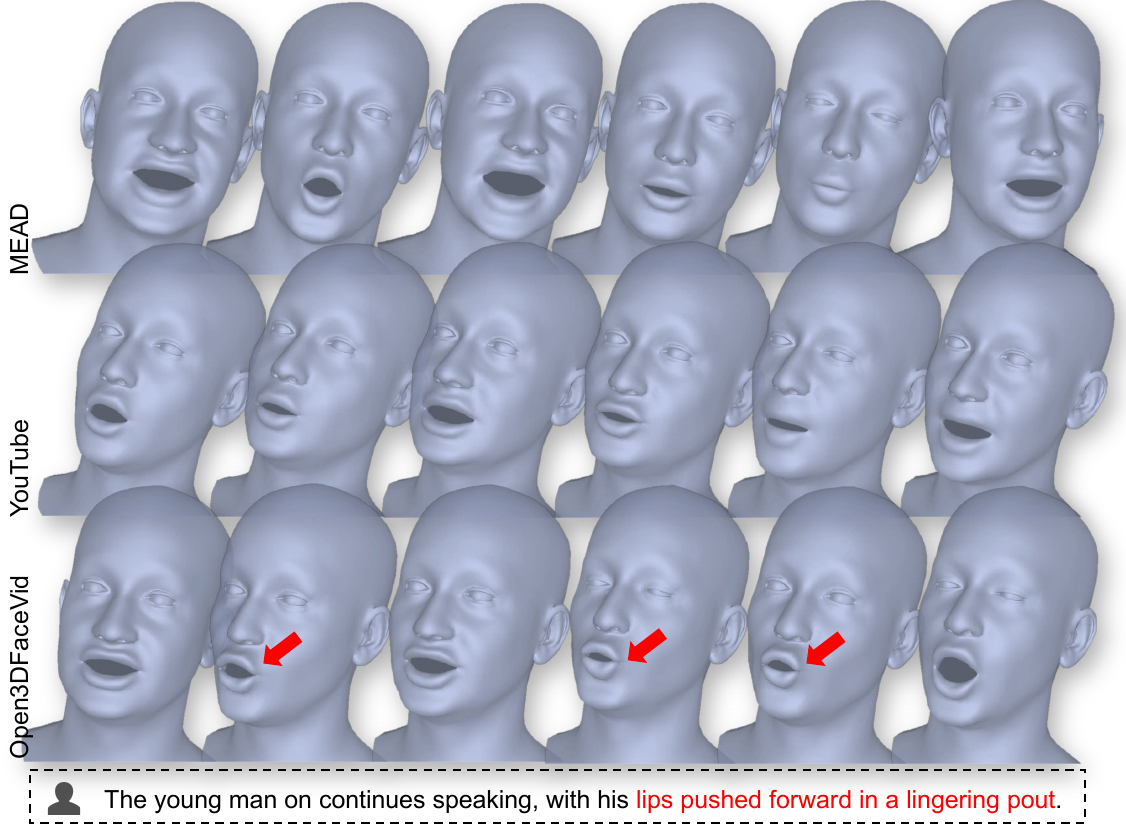}
\vspace{-0.5cm}
\caption{The qualitative ablation on training corpus for Language2Motion. We visualize facial motion generated by models trained on ablation datasets. The MEAD- and YouTube-trained models miss the \textit{“pout”} behavior, whereas the our model produces clear lip protrusion (red arrows), matching the description.}
\vspace{-0.4cm}
\label{fig:abliation-l2m-result-1}
\end{figure}

\begin{figure}[]
\centering
  \includegraphics[width=1\columnwidth, trim={0cm 0cm 0cm 0cm}, clip]{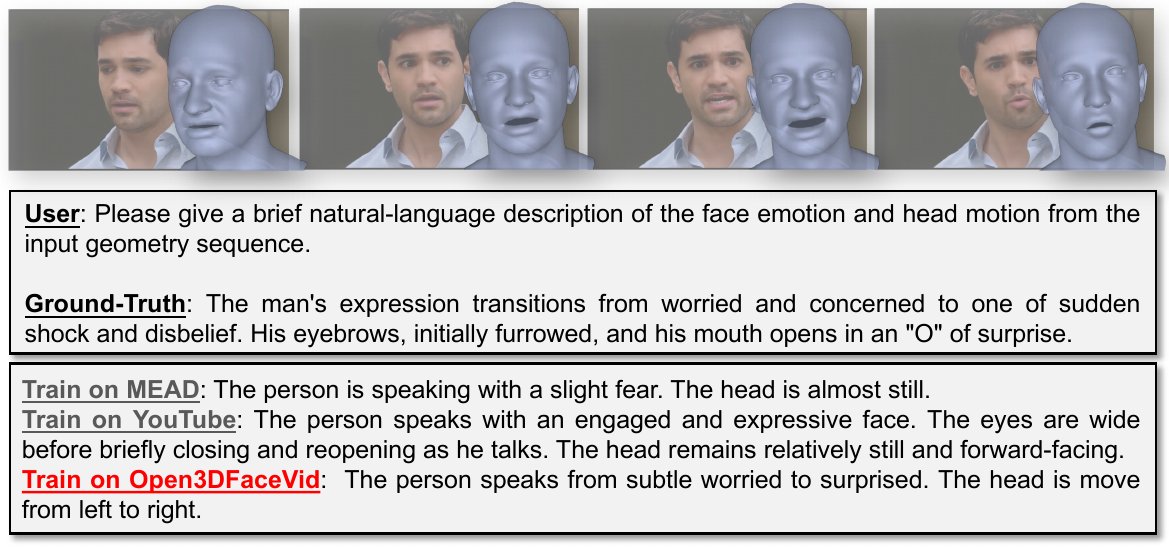}
\vspace{-0.5cm}
\caption{Qualitative ablation on corpus for Motion2Language. Given user query, we show responses from models trained on MEAD, YouTube and Open3DFaceVid. The MEAD- and YouTube-trained variants produce generic or partially correct descriptions, while the Open3DFaceVid model most accurately captures the subtle worry-to-surprise transition and head motion.}
\vspace{-0.3cm}
\label{fig:ablation-l2m-result-1}
\end{figure}

\subsection{Ablation Study}

\vspace{0.1cm}

\begin{figure}[tp]
\centering
  \includegraphics[width=1\columnwidth, trim={0cm 0cm 0cm 0cm}, clip]{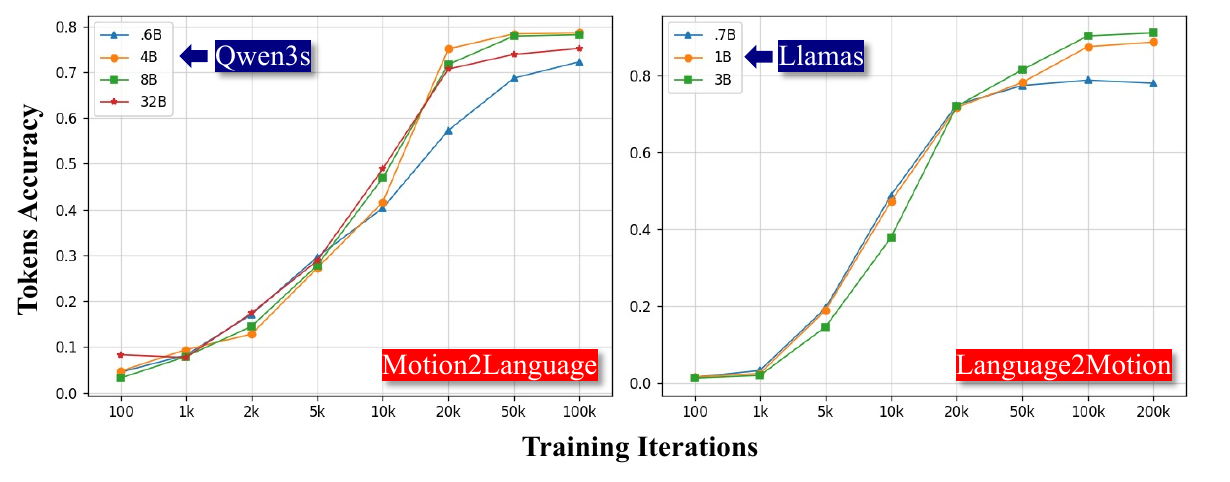}
\vspace{-0.8cm}
\caption{The scaling behavior of Motion2Language and Language2Motion. Left: token accuracy over training iterations for Motion2Language with Qwen3 backbones of different sizes (.6B-32B). Right: token accuracy for Language2Motion with LLaMA-based backbones (0.7B-3B).
}
\label{fig:ablation-scale-result-1}
\end{figure}

\begin{figure}[tp]
\centering
  \includegraphics[width=1.\columnwidth, trim={0cm 0cm 0cm 0cm}, clip]{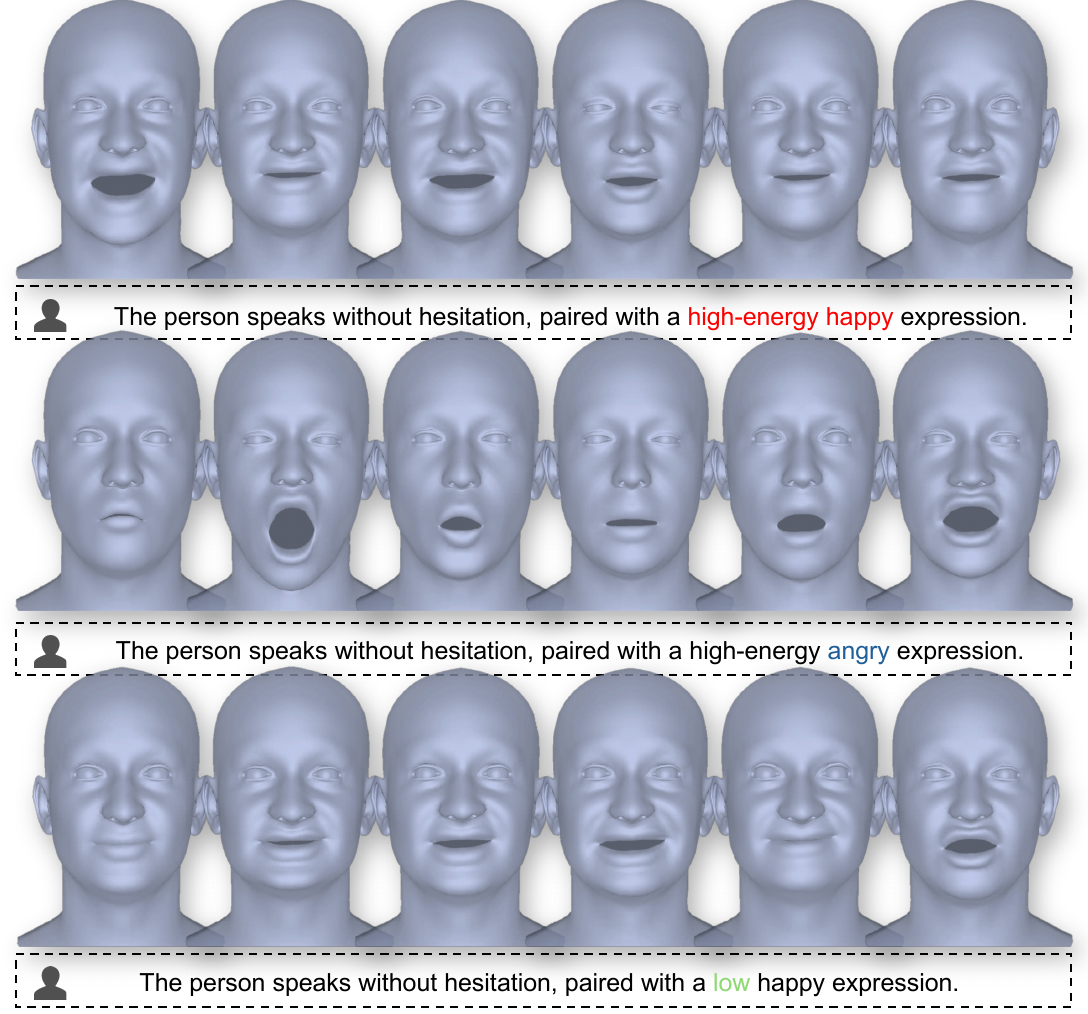}
\vspace{-0.5cm}
\caption{Language-controlled expressive ablation. We modify one keyword in prompt and let the Language2Motion model generate the corresponding facial motion. The two happy prompts produce different intensity level, while the angry prompt yields clearly distinct mouth shapes, demonstrating fine-grained text control over emotional style.}
\vspace{-0.3cm}
\label{fig:ablation-word-control}
\end{figure}

\noindent\textbf{Dataset Ablation.}
To understand the value of our synthetic corpus, we compare Open3DFaceVid against two representative alternatives:  
(1) the MEAD~\cite{wang2020mead} studio-captured dataset with one-hot emotion labels, converted into textual prompts; and  
(2) an in-the-wild YouTube set automatically annotated by Gemini.  
These two sources reflect the two dominant data paradigms in facial-behavior research, controlled lab capture and unconstrained internet videos, both of which are known to lack the fine-grained expressive coverage required for language-driven modeling.

\noindent  \textit{Evaluation Protocol.}  
We follow a TMR-style~\cite{petrovich2023tmr} protocol to assess whether each dataset provides semantically reliable text–motion pairs.  
For every dataset, we train a CLIP-like text–motion encoder on its paired annotations and project all samples into a shared embedding space.  
If a dataset contains high-quality, well-aligned annotations and exhibits sufficient expression diversity, its embeddings should form \emph{well-separated clusters} corresponding to different facial categories.  
Conversely, datasets with limited variation or noisy annotations should collapse into overlapping or ambiguous regions.

The t-SNE visualizations in Fig.~\ref{fig:tsne} confirm this behavior.  
Open3DFaceVid displays clear and balanced clusters across 187 labels, reflecting both the richness of our curated lexicon and the reliability of the T2V generation pipeline in following prompts.  
MEAD, despite being high-quality capture, offers only eight coarse labels and limited expressive variation, leading to tighter, less informative clusters.  
YouTube data, even after Gemini annotation, remains heavily skewed toward neutral expressions and exhibits considerable overlap, highlighting the difficulty of recovering subtle facial behaviors from uncontrolled videos.

\noindent \textit{Downstream comparisons.}  
We next train both Motion2Language and Language2Motion models on each dataset.  
For Motion2Language, the broader coverage of Open3DFaceVid yields noticeably richer semantic understanding (Fig.~\ref{fig:ablation-l2m-result-1}), even though its head-motion range is slightly narrower than the YouTube variant (reflected by $\text{Cor}_M$ in Tab.~\ref{table_4}).  
For Language2Motion, models trained on our dataset produce clearer expression articulation, the lingering \emph{pout} in Fig.~\ref{fig:abliation-l2m-result-1} for instance, demonstrating stronger text–motion alignment.  
Although MEAD occasionally shows lower feature-space distances due to its simpler label space, models trained on Open3DFaceVid achieve the highest USER scores (Tab.~\ref{table_5}), indicating superior perceptual quality and semantic fidelity.

Overall, these ablations show that neither controlled studio capture nor in-the-wild videos offer the diversity or balance needed for expressive facial-motion learning.  
Our synthetic corpus provides significantly stronger annotation quality, more comprehensive expressive coverage, and more reliable text–motion alignment, benefiting both Motion2Language and Language2Motion tasks.  
Additional analysis of dataset bias and synthetic-data effectiveness is provided in the Appendix.

\vspace{0.1cm}
\noindent \textbf{Scaling Behavior.} We present the scaling behavior by plotting token accuracy for Motion2Language and Language2Motion in Fig.~\ref{fig:ablation-scale-result-1}. For Motion2Language, performance saturates around 4B–8B parameters, and larger backbones offer limited gains. In contrast, Language2Motion benefits more strongly from scale, with larger models yielding higher token-generation accuracy.

\vspace{0.1cm}
\noindent \textbf{Language2Motion Keywords.} As shown in Fig.~\ref{fig:ablation-word-control}, we obtain facial motions by changing only a single word in the prompt. The edit can adjust style intensity (\textit{``high-energy"-``low"}) or swap the affective category itself (\textit{``happy"-``angry"}). These results demonstrate both the model nuanced language understanding and its ability to translate fine-grained prompts into controllable facial motion.

\vspace{-0.1cm}
\section{Conclusion}
\label{sec:conclusion}
\vspace{-0.1cm}

We study a previously underexplored direction, aligning language with 3D facial motion. We first analyze a key bottleneck, the scarcity of facial datasets with paired text and address it by synthesizing a large-scale corpus of facial videos with T2V models. Building on this resource, we study two complementary settings: Motion2Language and Language2Motion. We hope our dataset and bidirectional framework will serve as a foundation for future research in 3D facial motion understanding and animation.

{
    \small
    \bibliographystyle{ieeenat_fullname}
    \bibliography{main}
}


\end{document}